\documentclass{esannV2}
\usepackage[dvips]{graphicx}
\usepackage[latin1]{inputenc}
\usepackage{amssymb,amsmath,array}
\usepackage{subfig}
\usepackage{bm}
\newcommand{\x}{\mathbf{x}}
\renewcommand{\u}{\mathbf{u}}

\newcommand{\y}{\mathbf{y}}
\newcommand{\W}{\mathbf{W}}

\newcommand{\Wh}{\hat{\W}}

\newcommand{\R}{\mathbb{R}}

\newcommand{\xl}[1]{\x^{(#1)}}

\newcommand{\Fl}[1]{F^{(#1)}}

\begin{document}

\title{Short-term Memory of Deep RNN}

\author{Claudio Gallicchio
\vspace{.3cm}\\
Department of Computer Science, University of Pisa\\
Largo Bruno Pontecorvo 3 - 56127 Pisa, Italy\\
}

\maketitle

\begin{abstract}
The extension of deep learning towards temporal data processing is gaining an increasing research interest.
In this paper we investigate the properties of state dynamics developed in successive levels
of deep recurrent neural networks (RNNs) in terms of short-term memory abilities.
Our results reveal interesting insights that shed light on the nature of layering as a factor of RNN design.
Noticeably, higher layers in a hierarchically organized RNN architecture results to be inherently biased
towards longer memory spans even prior to training of the recurrent connections.
Moreover, in the context of Reservoir Computing framework, our analysis also points out the benefit
of a layered recurrent organization as an efficient approach to improve the memory skills of reservoir models.
\end{abstract}

\section{Introduction}
Deep learning is an attractive area of research in constant growth \cite{DeepLearningBook}.
In particular, in the neuro-computing field, the study of deep neural networks
composed by multiple non-linear layers has proved able to learn feature representations
at progressively higher levels of abstraction, leading to eminent performance e.g. in vision tasks.
Extending the benefits of depth to recurrent neural networks (RNNs) is
an intriguing research direction that is recently gaining an increasing attention \cite{Angelov2016}.
In this context, the study of deep RNNs has pointed out that hierarchically organized recurrent models 
have the potentiality of developing multiple time-scales 
representations of the input history in their internal states, which can be of great help, e.g.,
when dealing with
text processing tasks \cite{Hermans2013}.
More recently, studies in the area of Reservoir Computing (RC)  \cite{Verstraeten2007, Lukosevicius2009}
have  shown that the ability of 
developing such a structured state space organization 
is indeed an intrinsic property of  layered RNN architectures \cite{Gallicchio2017DeepESN, Gallicchio2017echo}.
The study of deep RC networks on the one hand allowed the development of efficiently trained deep models for learning in the temporal domain, and on the other hand it paved the way to further studies on the properties  
of deep RNNs dynamics even in the absence of (or prior to) learning
of the recurrent connections.

An aspect of prominent relevance in the study of dynamical models is represented by 
the analysis of their memory abilities.
In this paper, exploiting the ground provided by the deep RC framework, we explicitly 
address the problem of analyzing the short-term memory capacity of individual (progressively higher) layers in deep recurrent architectures.
Contributing to  highlight the
intrinsic diversification of transient state dynamics in hierarchically 
constructed recurrent networks, our investigation aims at shedding more light on the 
bias of layering in the RNN architectural design.
Framed in the RC area, our analysis is also intended to provide insights on 
the process of reservoir network construction. 


\section{Deep Stacked RNN}
\label{sec.deepRNN}
\begin{figure}[tbh]
	\centering
		\includegraphics[width=0.85\textwidth]{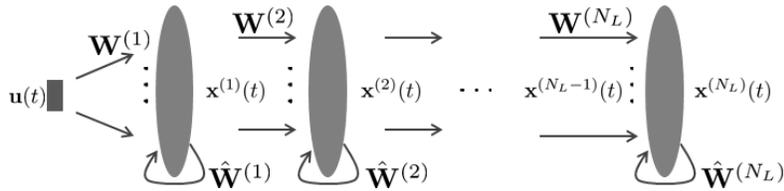}
	\caption{Hierarchical organization of hidden layers in a deep RNN.}
	\label{fig.architecture}
\end{figure}

We consider deep RNNs \cite{Hermans2013} whose recurrent architecture is obtained by a stacked composition of
multiple non-linear recurrent hidden layers, as illustrated in Fig.~\ref{fig.architecture}.
The state computation proceeds by following the hierarchical network organization, from the lowest layer to the highest one.
Specifically, at each time step $t$ the first recurrent layer in the network is fed by the external input 
while each successive layer is fed by the activation of the previous one.

Under a dynamical system perspective, 
a deep RNN implements an input-driven discrete-time non-linear dynamical system,
in which the state evolution in each layer $i$ is ruled by 
a state transition function 
$\Fl{i}$.
Here we denote the input dimension by $N_U$ and assume, for the sake of simplicity,  that each hidden layer contains $N_R$ recurrent units. In the following, we use $\u(t)$ and $\xl{i}(t)$, respectively, to indicate the external input and state of i-th hidden layer at time step $t$. 
Based on this notation, the state in the first layer is updated according to the following equation:
\begin{equation}
\label{eq.layer1}
\x^{(1)}(t) = \Fl{1}(\u(t), \xl{1}(t-1)) = 
\tanh(\W^{(1)} \u(t) + \Wh^{(1)} \xl{1}(t-1)),
\end{equation}
where $\W^{(1)} \in \R^{N_R \times N_U}$ is the input weight matrix and $\Wh^{(1)} \in \R^{N_R \times N_R}$ is the
recurrent weight matrix for the first layer.
For each successive layer $i>1$, the state is updated according to:
\begin{equation}
\label{eq.layeri}
\xl{i}(t) = \Fl{i}(\xl{i-1}(t), \xl{i}(t-1)) = 
\tanh(\W^{(i)} \xl{i-1}(t) + \Wh^{(i)} \xl{i}(t-1)),
\end{equation}
where $\W^{(i)} \in \R^{N_R \times N_R}$ collects the weights for the inter-layer connections from layer $i-1$ to layer $i$ and 
$\Wh^{(i)} \in \R^{N_R \times N_R}$ is the recurrent weight matrix for layer $i$.
Note that in both above eq. \ref{eq.layer1} and \ref{eq.layeri}, 
a $\tanh$ non-linearity is used as element-wise applied activation function of recurrent units and
bias terms are omitted for the ease of notation.
Here it is also worth observing that,
although the deep recurrent dynamics globally evolve 
as a whole, 
locally to each layer $i$ the state information coming from the previous level $i-1$
actually acts as an independent input information that encodes the history of the external input
up to the present time step. 

Taking aside the aspects related to learning of the recurrent connections (and the specific aspects involved by different training strategies), here we focus our analysis 
on the case of untrained deep recurrent dynamics. To do so, we resort to the recently 
introduced deep RC framework \cite{Gallicchio2017DeepESN}, according to which the recurrent part of the deep RNN architecture is left untrained after initialization subject to stability constraints \cite{Gallicchio2017echo}. Specifically, the network is initialized 
with weights  from a uniform distribution in $[-1,1]$ and then re-scaled to control
for each layer $i$ the values of 
$\|\W^{(i)}\|_2$ and $\rho(\Wh^{(i)})$, where $\rho(\cdot)$ denotes the spectral radius of its matrix argument (i.e. the maximum among the eigenvalues magnitudes). These quantities are
hyper-parameters of the model that influence its state dynamics
and that are typically set to small values in order to guarantee a stable regime,
a standard initialization approach also for trained networks.
Notice that this framework
allows us on the one hand to investigate the fixed characterization
of state dynamics in successive levels of a deep RC network, and on the other hand 
to study of the bias due to layering in deep recurrent architectures.

Output computation is 
implemented by using an output layer of size $N_Y$.
Though different choices are possible 
for the state-output connection settings
(see e.g. \cite{Hermans2013,Pascanu2014}), following from our analysis aims here we consider output modules that are individually applied to each layer of the recurrent network. This enables us to study separately the 
characteristics of
the state behavior emerging  at the different levels in the architecture.
We use linear output modules, such that for each layer $i$ the output is computed as
$\y^{(i)}(t) = \W_{out}^{(i)} \xl{i}(t)$, where matrices $\W_{out}^{(i)} \in \R^{N_Y \times N_R}$
are trained for each layer individually, using a direct method such as pseudo-inversion.
In the RC framework this setting also ensures the same training cost
for every layer.

\section{Experiments}
We investigate the short-term memory abilities
of deep RNN architectures by resorting to the
Memory Capacity (MC) task \cite{Jaeger2001short}.
This aims at measuring 
the extent to which past input events can be recalled from present
state activations. 
Specifically, the recurrent system is tested in its ability to 
reconstruct delayed versions of a stationary uni-variate 
driving input signal ($N_U = N_Y = 1$),
with the MC at layer $i$ computed as a squared correlation coefficient, as follows:
\begin{equation}
\label{eq.mc}
MC^{(i)} = \sum_{k=1}^\infty MC_k^{(i)} =
\sum_{k=1}^\infty \frac{Cov^2(u(t-k),y_k^{(i)}(t))}{Var(u(t))\, Var(y_k^{(i)}(t))},
\end{equation}
where $y_k^{(i)}(t)$ is the activation of the output unit trained to reconstruct the $u(t-k)$ signal
from the state of layer $i$, while $Cov$ and $Var$ respectively denote the covariance and variance operators.
In order to maximally exercise the memory capability of the systems, we used i.i.d. input signals
from a uniform distribution in $[-0.8,0.8]$, i.e. an unstructured temporal stream in which $u(t)$
does not carry information on previous inputs $\ldots, u(t-2), u(t-1)$.
For this task, we considered a $6000$ time-step long sequence, where the first $5000$ time steps
were used for training\footnote{The first 1000 time steps were 
considered as transient to washout the initial conditions.}
and the remaining $1000$ for MC assessment.

We considered deep RNNs with $N_L = 10$ recurrent layers, each of which
containing $N_R = 100$ recurrent units. 
Input and inter-layer weights were re-scaled such that 
$\|\W^{(i)}\|_2 = 1$ for $i = 1,\ldots,N_L$. Weights of recurrent connections were
re-scaled to the same spectral radius in all the layers, i.e. $\rho = \rho(\Wh^{(i)})$ for $i =1,\ldots,N_L$, with $\rho$ values ranging in $[0.1, 1.5]$. 
Note that, under the considered experimental settings, for higher values of $\rho > 1$ 
the network dynamics tend to exhibit a chaotic behavior, as shown in previous works
in terms of local Lyapunov exponents 
\cite{Gallicchio2017ESANN,Gallicchio2018Lyapunov}.
Although recurrent dynamics in chaotic regimes are generally not interesting under a practical point of view,
in this paper we consider also these cases ($\rho > 1$) for the scope of analysis.
For each choice of $\rho$ we independently generated $50$ networks realizations (with different seeds for random generation), and averaged the achieved results over such realizations. 

For practical assessment of the MC values, it is useful to recall a basic 
theoretical result provided in \cite{Jaeger2001short}, which states that the MC of an $N_R$-dimensional
recurrent system driven by an i.i.d. uni-variate input signal is upper bounded by $N_R$.
Accordingly, we considered a maximum value for the delay $k$ in eq.~\ref{eq.mc}
equal to twice the size of the state space, i.e. $200$,
which is sufficient to  account for the correlations that are practically involved
in our experimental settings.

\begin{figure}[b!]
	\centering
		\includegraphics[width=0.8\textwidth]{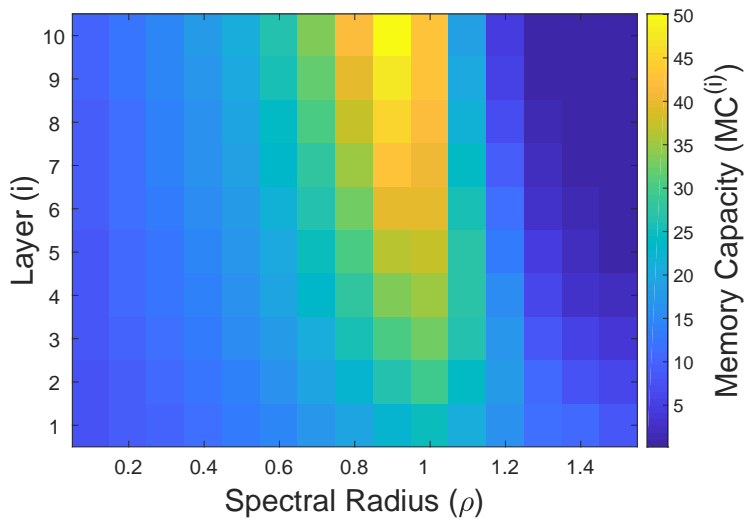}
	\caption{MC of different layers in deep RNNs for increasing values of $\rho$.}
	\label{fig.mc}
\end{figure}

Fig. \ref{fig.mc}
shows the MC values achieved in correspondence of progressively higher layers in the architecture,
and for the different cases of $\rho$ considered for network initialization.
Results clearly point out that for recurrent networks in the ordered regime ($\rho$ not exceeding 1) 
higher layers in the deep architecture 
are naturally biased toward 
progressively longer 
short-term memory abilities. 
For networks in a chaotic regime ($\rho$ above 1) higher layers tend to show a poorer MC.
The MC performance shown in Fig.~\ref{fig.mc} has a peak in correspondence of 
$\rho = 0.9$, 
in which case the score improves from $22.2$ in the 1st layer, up to $50.1$ in the 10th layer.
Interestingly, our results also highlight the effectiveness of layering
and its striking advantage 
as a convenient process for RC networks architectural
design. 
The memory of an $N_R$-dimensional 
reservoir can be indeed easily improved
by using an underlying stack of recurrent layers to filter the external input signal.
Note that such improvement comes at the only  price of a modestly increased cost for the state computation
(that increases linearly with the number of layers), while the 
cost for output training remains the same.

\begin{figure}[bth]
	\centering
	\vspace{0.15cm}
		\includegraphics[width=0.8\textwidth]{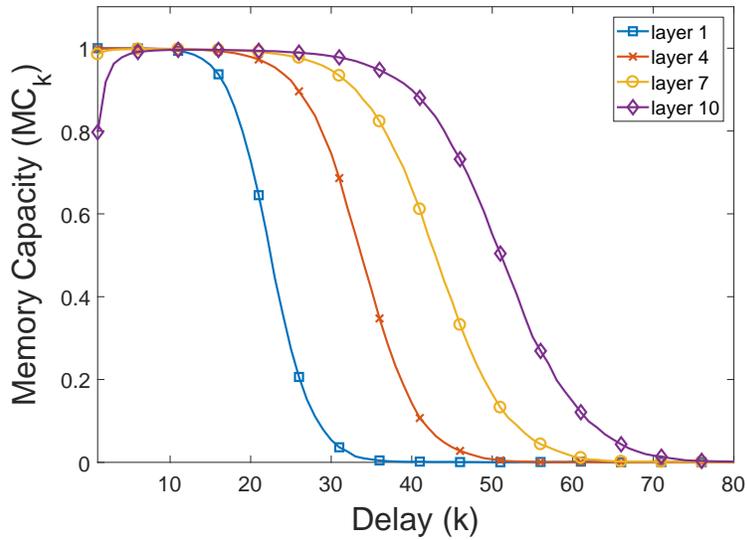}
	\caption{k-delay memory capacity of deep RNN layers at increasing height.}
	\label{fig.fc}
\end{figure}

We further inquire into the memory structure developed by the layers of deep RNNs by analyzing the MC values for
increasing delays. Fig.~\ref{fig.fc} shows the forgetting curves of individual (progressively higher) layers,
i.e. the values of $MC_k^{(i)}$ as a function of $k$, obtained in the case of $\rho = 0.9$.
The plot in Fig.~\ref{fig.fc} clearly reveals the diversification of 
memory spans in the components of the deep recurrent architecture: 
higher layers are able to store information about past inputs for longer times.
While for the 1st layer the memory recall is almost null after delay $30$, 
the dynamics developed in the 10th layer lead to a value that is 
above zero even for a delay of $60$.
We can also see that input signals with smaller delays are better reconstructed in the lower layers,
while higher layers are characterized by a peak that tends to shift to the right (more evident for layer 10 in Fig.~\ref{fig.fc}), and by a slope of the forgetting curve that tends to be increasingly smoother.
Besides, the highlighted diversification of short-term memory spans among the successive layers in the 
deep RNN architecture is also interesting as a way of characterizing (in a quantitative way) the intrinsic richness 
of  state representations 
globally developed by the deep recurrent system.

\section{Conclusions}
In this paper we have provided a computational analysis of short-term memory in deep RNNs.
To do so, we have resorted to the MC task as a mean to quantify the memory 
of state dynamics in successive levels of a deep recurrent system.

Our results clearly showed that higher layers in a hierarchically organized RNN architecture are
inherently featured, even prior to learning of recurrent connections, 
by an improved ability to latch input information for longer time spans.
The analysis provided in this paper 
also revealed interesting insights on the diversification of the memory structure developed 
within deep stacked RNN dynamics,
showing that higher layers tend to forget the past input history more slowly and more smoothly
compared lower ones. Furthermore, framed within the deep RC framework,
our results provided evidence that 
support the practical benefit of the layered recurrent
organization as a way to improve the memory skills of reservoir networks
in a cost-effective fashion.

Overall, though further studies in this research direction are certainly demanded
(e.g. on the theoretical side),
we believe that the outcomes provided in this paper can contribute 
to better understand and characterize the bias due to layering
in deep recurrent neural models.


\begin{footnotesize}


\bibliographystyle{unsrt}
\bibliography{references}

\end{footnotesize}


\end{document}